\documentclass[nohyperref]{article}

\usepackage{microtype}
\usepackage{graphicx}
\usepackage{booktabs} 

\usepackage[accepted]{icml2023}

\usepackage{amsmath}
\usepackage{amssymb}
\usepackage{mathtools}
\usepackage{amsthm}

\theoremstyle{plain}

\theoremstyle{definition}

\theoremstyle{remark}

\usepackage[textsize=tiny]{todonotes}

\usepackage{amsmath}
\usepackage[scaled=0.92]{PTSans}
\usepackage{xfrac}

\usepackage[utf8]{inputenc} 
\usepackage[T1]{fontenc}    
\usepackage[colorlinks]{hyperref}
\hypersetup{citecolor=MidnightBlue}
\hypersetup{linkcolor=black}
\hypersetup{urlcolor=MidnightBlue}
\usepackage{url}            
\usepackage{booktabs}       
\usepackage{nicefrac}       
\usepackage{microtype}      

\usepackage{graphicx} 

\usepackage{amssymb}
\usepackage{pifont}
\usepackage{graphicx}
\usepackage{caption}
\usepackage{subcaption}
\usepackage{lipsum}
\usepackage{bbm}
\usepackage{cancel}
\usepackage{paralist}
\usepackage[inline]{enumitem}
\usepackage{multirow}
\usepackage{array}

\usepackage{changepage}
\usepackage{adjustbox}

\usepackage{amsfonts}
\usepackage{amsmath}
\usepackage{amssymb}
\usepackage{amsthm}

\usepackage{listings}
\usepackage{xcolor}
\usepackage{courier}

\definecolor{codegreen}{rgb}{0,0.6,0}
\definecolor{codegray}{rgb}{0.5,0.5,0.5}
\definecolor{codepurple}{rgb}{0.58,0,0.82}
\definecolor{backcolour}{rgb}{0.95,0.95,0.92}

\lstdefinestyle{mystyle}{
    backgroundcolor=\color{backcolour},   
    commentstyle=\color{codegreen},
    keywordstyle=\color{magenta},
    numberstyle=\tiny\color{codegray},
    stringstyle=\color{codepurple},
    basicstyle=\small\ttfamily,
    breakatwhitespace=false,         
    breaklines=true,                 
    captionpos=b,                    
    keepspaces=true,                 
    numbers=left,                    
    numbersep=5pt,                  
    showspaces=false,                
    showstringspaces=false,
    showtabs=false,                  
    tabsize=2
}

\usepackage{todonotes}

\usepackage{multicol}
\usepackage{soul}
\newcommand{\N}{\mathcal{N}}

\newcommand{\R}{\mathbb{R}}

\newcommand{\KL}{\text{KL}}

\DeclareMathOperator*{\argmin}{arg\,min}

\icmltitlerunning{Revisiting Structured VAEs}

\begin{document}

\twocolumn[
\icmltitle{Revisiting Structured Variational Autoencoders}

\begin{icmlauthorlist}
\icmlauthor{Yixiu Zhao}{xxx}
\icmlauthor{Scott W. Linderman}{yyy}
\end{icmlauthorlist}

\icmlaffiliation{xxx}{Applied Physics Department, Stanford University}
\icmlaffiliation{yyy}{Department of Statistics and the Wu Tsai Neurosciences Institute, Stanford University}
\icmlcorrespondingauthor{Yixiu Zhao}{yixiuz@stanford.edu}

\icmlkeywords{Machine Learning, Bayesian Inference, ICML}

\vskip 0.3in
]

\printAffiliationsAndNotice{}  

\begin{abstract}

Structured variational autoencoders (SVAEs) \citep{johnson2016composing} combine probabilistic graphical model priors on latent variables, deep neural networks to link latent variables to observed data, and structure-exploiting algorithms for approximate posterior inference. 
These models are particularly appealing for sequential data, where the prior can capture temporal dependencies.
However, despite their conceptual elegance, SVAEs have proven difficult to implement, and more general approaches have been favored in practice. 
Here, we revisit SVAEs using modern machine learning tools and demonstrate their advantages over more general alternatives in terms of both accuracy and efficiency. 
First, we develop a modern implementation for hardware acceleration, parallelization, and automatic differentiation of the message passing algorithms at the core of the SVAE.
Second, we show that by exploiting structure in the prior, the SVAE learns more accurate models and posterior distributions, which translate into improved performance on prediction tasks. 
Third, we show how the SVAE can naturally handle missing data, and we leverage this ability to develop a novel, self-supervised training approach.
Altogether, these results show that the time is ripe to revisit structured variational autoencoders.
\end{abstract}

\section{Introduction}
\label{section:intro}

Variational autoencoders~\citep[VAEs,][]{kingma2013auto,rezende2014stochastic} are deep generative models that use neural networks to link latent variables to high dimensional observations.
\emph{Structured} variational autoencoders (SVAEs)~\citep{johnson2016composing} use probabilistic graphical models to capture dependencies in the prior distribution over latent variables.
For example, when working with sequential data, graphical models can capture latent dynamics~\citep{krishnan2015deep,archer2015black}.
When domain knowledge is available, graphical models offer an expressive means of incorporating it~\citep{pandarinath2018inferring,lopez2018deep}. 
When inferring the posterior distribution over latent variables, graphical models offer an ancillary benefit: SVAEs can leverage the rich toolkit of message passing algorithms for graphical models~\citep{wainwright2008graphical} to aid in approximate inference.

Specifically, the SVAE leverages a structure-exploiting amortized variational inference algorithm.
Rather than producing a full posterior distribution over latent variables, the recognition network (also referred to as an encoder) outputs \textit{conjugate potentials}.
When the prior is composed of exponential family distributions, the conjugate potentials can be combined with the prior using message passing algorithms to obtain an approximate posterior.
Essentially, the SVAE learns to approximate complex deep generative models with simpler exponential family models where fast and efficient algorithms can be applied.

Despite their conceptual elegance, two issues have limited the adoption of SVAEs.
The first is that the structure-exploiting inference algorithm is difficult to implement in practice, requiring gradients through message passing algorithms and fixed-point iterations. 
The second is a more basic question: what is the value of structure-exploiting amortized inference when we can construct arbitrarily flexible recognition networks?
In practice, some methods retain a structured prior but use more general recognition networks~\citep{krishnan2015deep,dilokthanakul2016deep,casale2018gaussian}.
Other methods use more flexible priors as well; e.g., sequential latent variable models often leverage recurrent neural network priors~\citep[e.g.][]{chung2015recurrent, karl2016deep, buesing2018learning, pandarinath2018inferring, kosiorek2018sequential, hafner2019learning, saxena2021clockwork}

\begin{figure*}[ht!]
\centering
\hspace{2mm}
\begin{subfigure}{.19\textwidth}
  \centering
  \includegraphics[trim={9cm 5.5cm 9cm 5cm},clip,width=1.\linewidth]{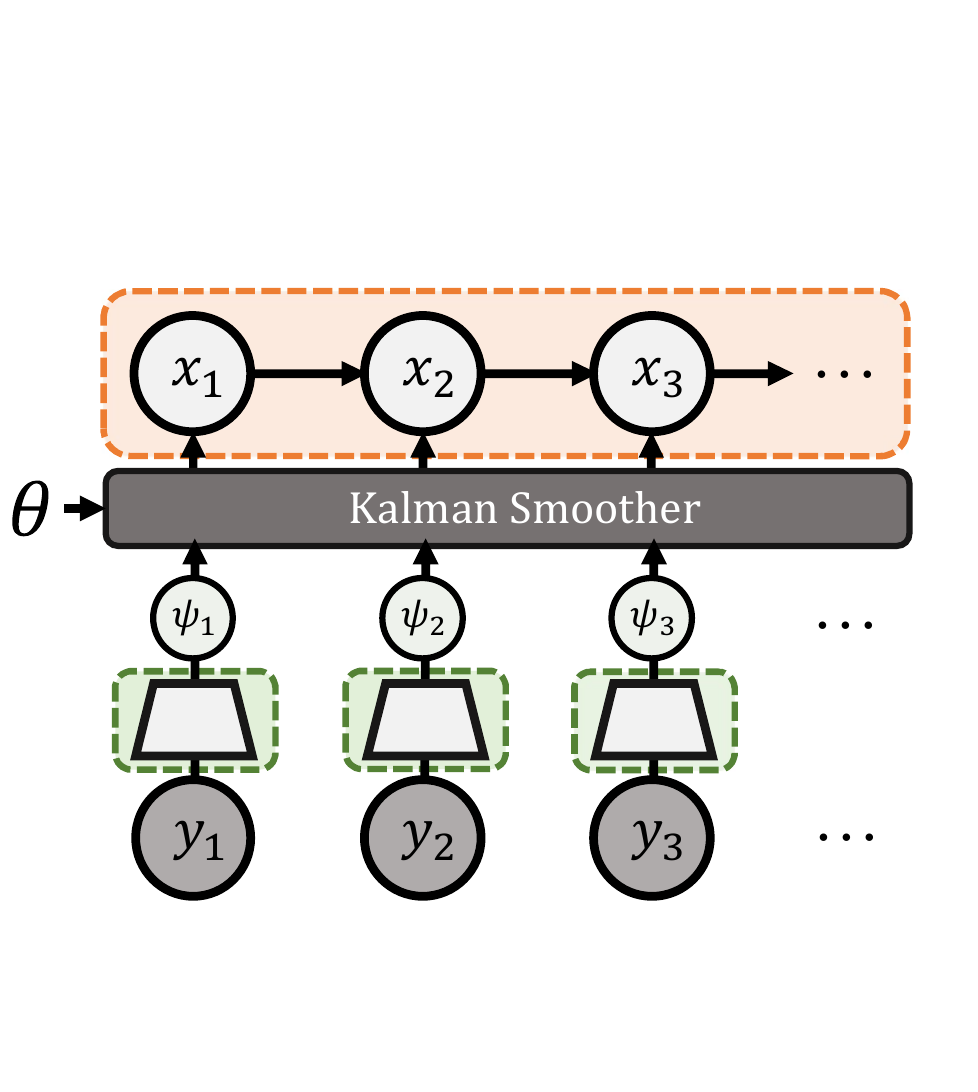}
  \vspace{-3em}
  \caption{SVAE}
  \label{fig:svae_inference}
\end{subfigure}
\begin{subfigure}{.19\textwidth}
  \centering
  \includegraphics[trim={8cm 5cm 8cm 5cm},clip,width=1.1\linewidth]{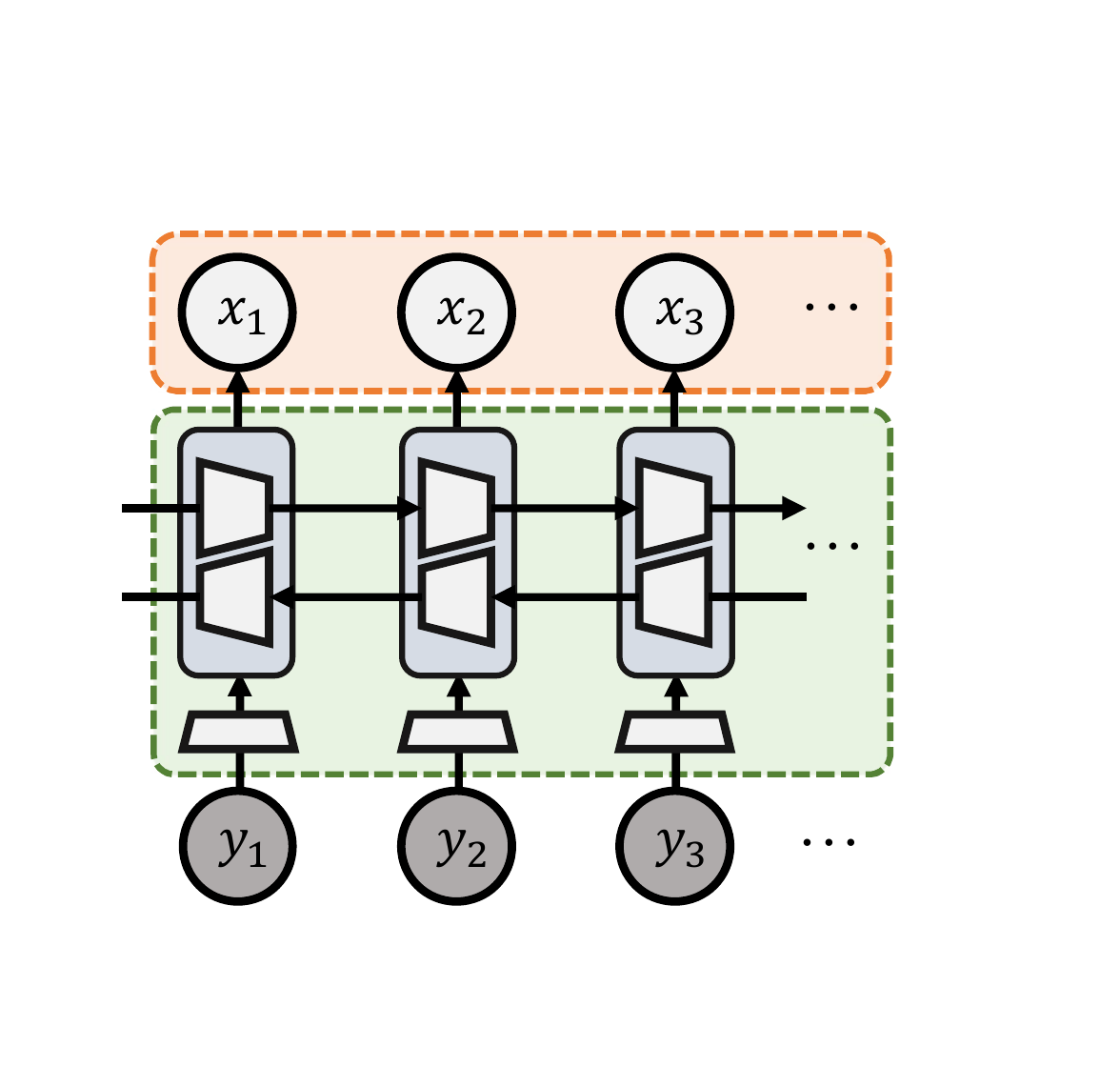}
  \vspace{-3em}
  \caption{RNN-MF}
  \label{fig:dkf_inference}
\end{subfigure}
\begin{subfigure}{.19\textwidth}
  \centering
  \includegraphics[trim={8cm 5cm 8cm 5cm},clip,width=1.1\linewidth]{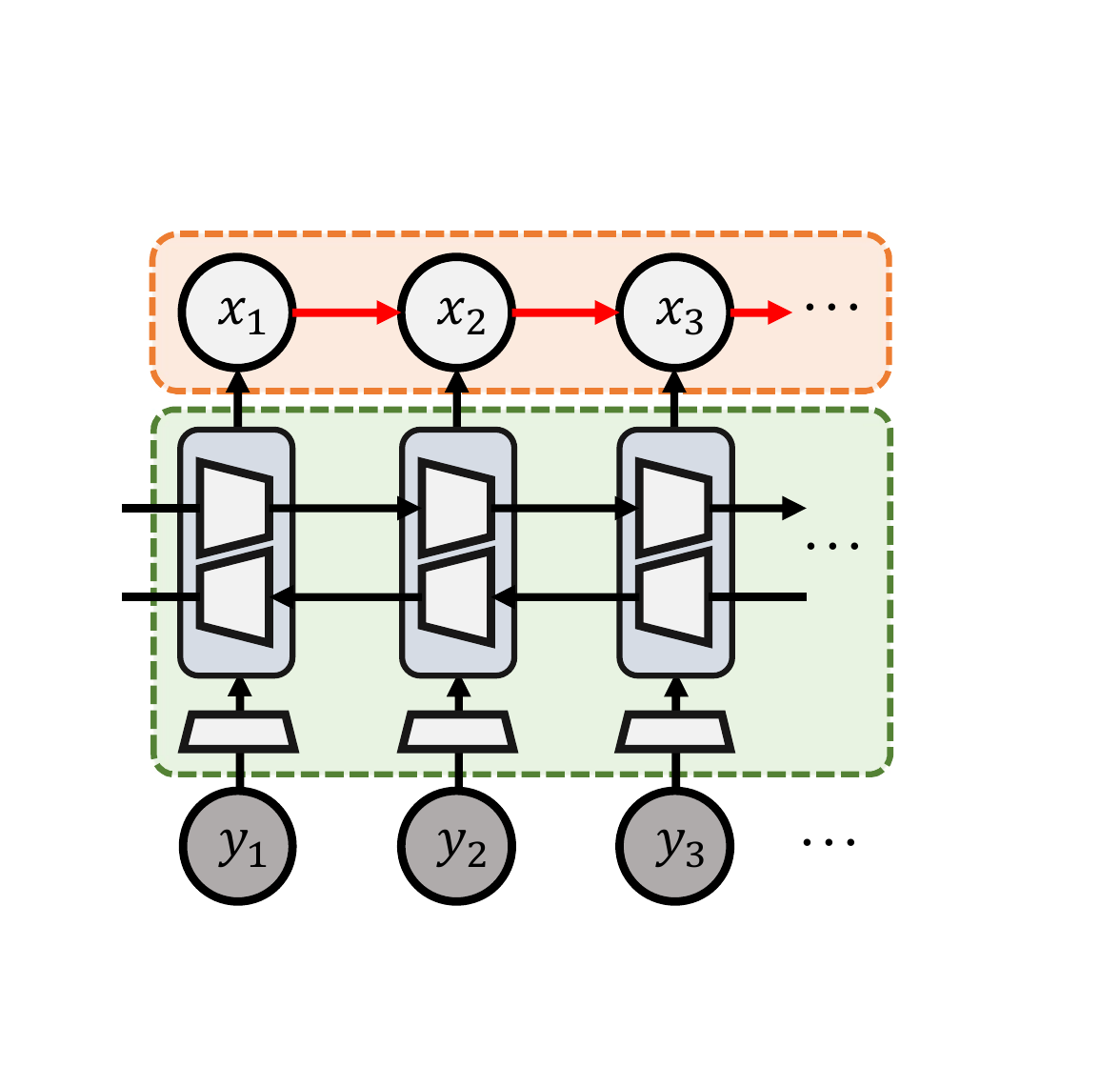}
  \vspace{-3em}
  \caption{RNN-AR-L}
  \label{fig:cdkf_inference}
\end{subfigure}
\begin{subfigure}{.19\textwidth}
\centering
  \includegraphics[trim={8cm 5cm 8cm 5cm},clip,width=1.1\linewidth]{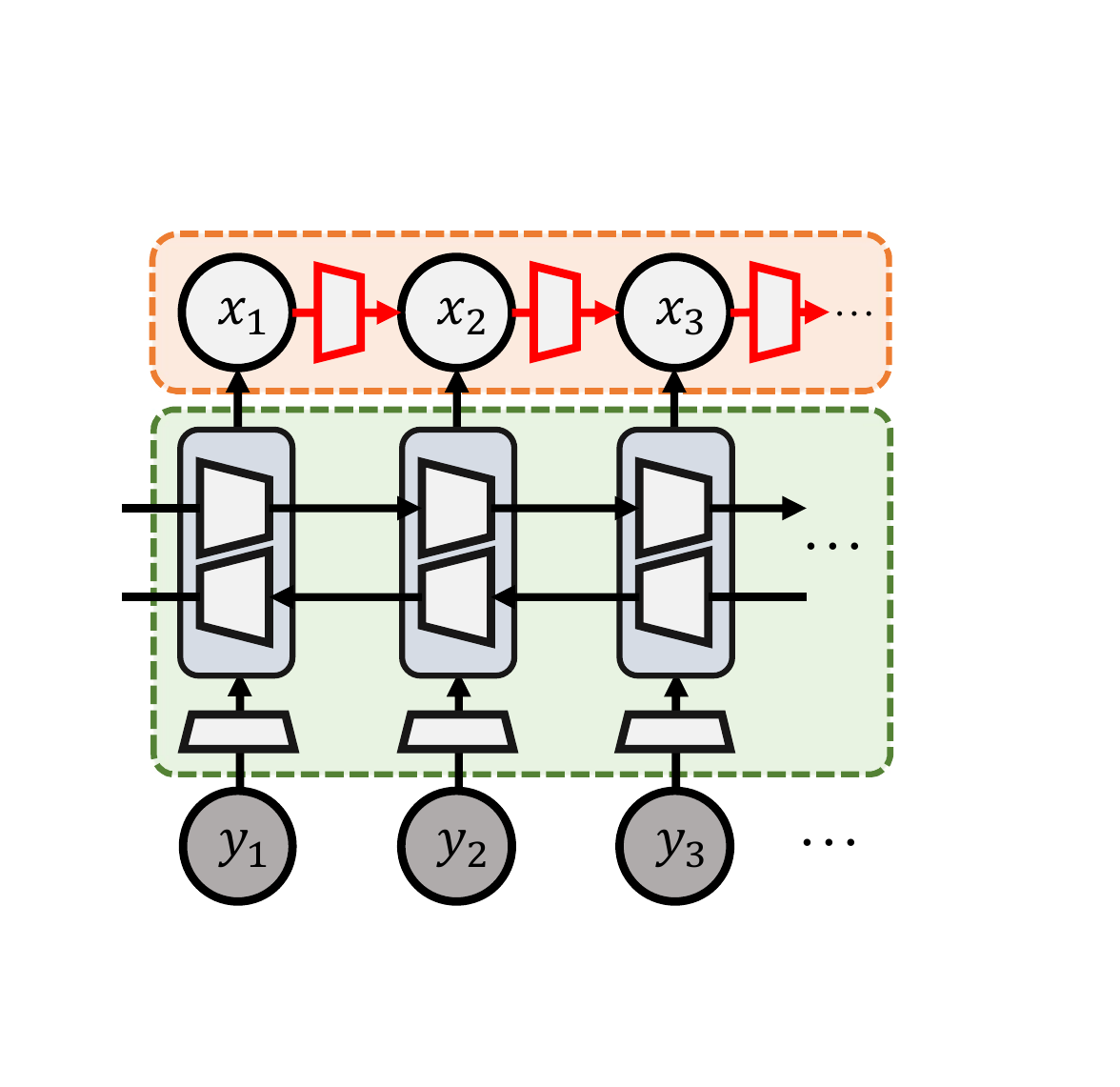}
  \vspace{-3em}
  \caption{RNN-AR-NL}
  \label{fig:planet_inference}
\end{subfigure}
\begin{subfigure}{.19\textwidth}
\centering
  \includegraphics[trim={8cm 5cm 8cm 5cm},clip,width=1.1\linewidth]{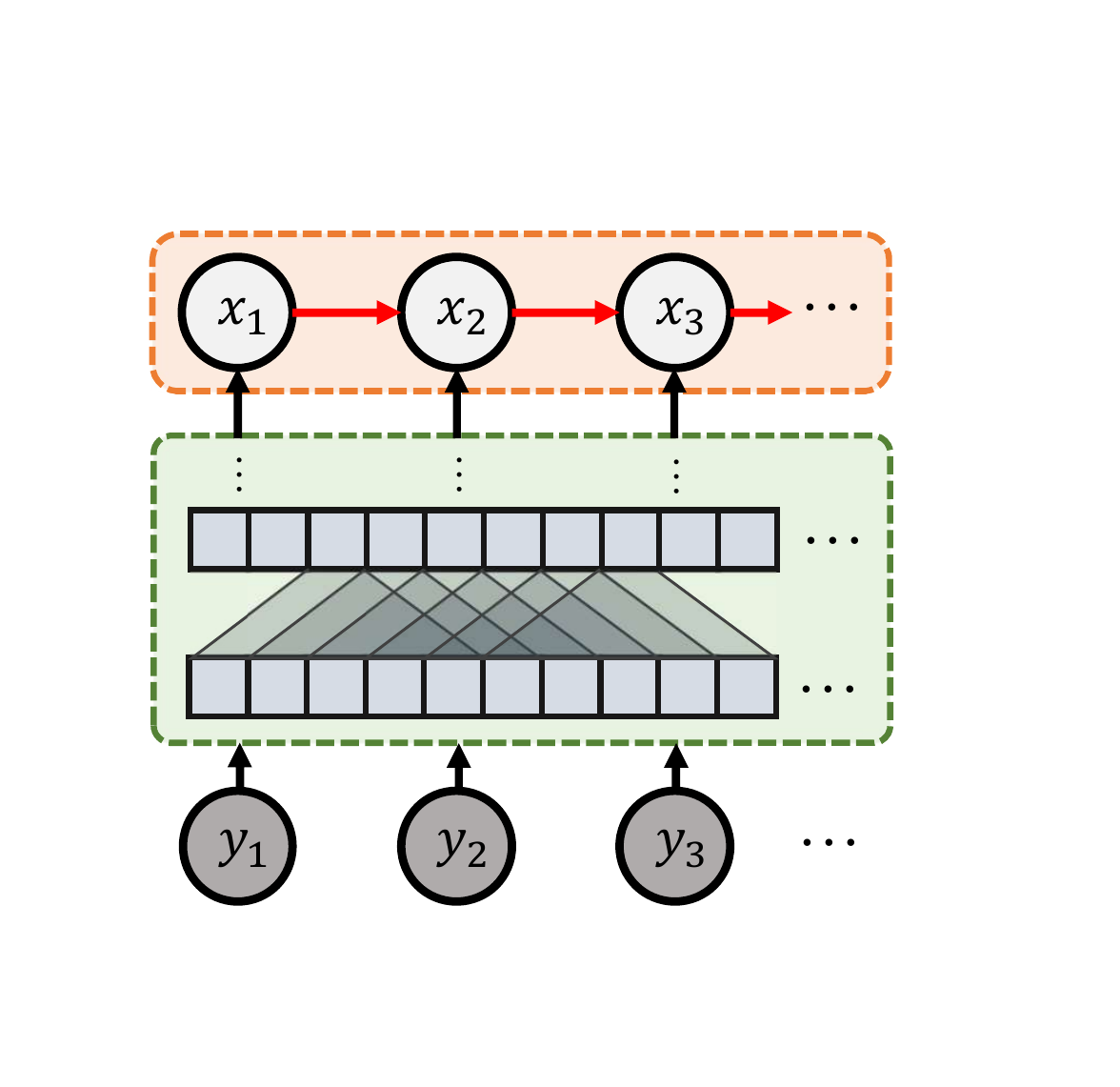}
  \vspace{-3em}
  \caption{CNN-AR-L}
  \label{fig:conv_inference}
\end{subfigure}
  \caption{Different recognition network architectures for sequential VAEs. (a) In the structured variational autoencoder~(SVAE), a neural network outputs recognition potentials $\psi_t$ for each time step, which are combined with the prior parameters $\theta$ via a message passing algorithm (e.g.~Kalman smoothing) to form the variational posterior $q(x_{1:T})$. Note that although $\psi_t$ only depends on $y_t$, the posterior has correlations across time due to message passing. 
  (b) In the mean field RNN recognition (RNN-MF) posterior family, a bi-directional RNN carries information forwards and backwards in time, generating a mean field variational posterior.
  (c-e) Temporal dependencies in the posterior can be captured with linear (c) or nonlinear (d) autoregressive dependencies that, again, depend on the outputs of a bi-directional RNN. This temporal dependency can also be approximated with a temporal CNN network with a finite kernel size (e).}
  \label{plot:linear_influence}
\end{figure*}

In this work we address both of these concerns. 
First, we develop a modern implementation of the SVAE that is fast and scalable with CPU, GPU, or TPU hardware.
We show how the SVAE can leverage parallel message passing algorithms to obtain orders of magnitude speedup over less structured alternatives.
Second, we study SVAEs for sequential data and demonstrate many advantages over more commonly used alternatives.
We show that structure-exploiting inference yields more accurate posterior approximations.
These improvements lead to better model learning and more accurate prediction, especially with higher dimensional latent states and in lower signal-to-noise regimes.
Finally, we highlight how SVAEs can naturally handle missing data, which motivates a novel, self-supervised training regimen to further improve model learning and predictive performance.
With a modern implementation and multiple improvements, it is time to revisit structured variational autoencoders.
%



\section{Background}
\label{section:background}
We begin with an introduction to structured variational autoencoders (SVAEs)~\citep{johnson2016composing}, with a particular emphasis on modeling sequential data. 
We then contrast the SVAE with more commonly used VAEs for sequential data, which use more general approaches to approximate posterior inference. 

\subsection{Structured Variational Autoencoders}
The SVAE~\cite{johnson2016composing} is both a modeling idea and an inference idea.
The modeling idea has been independently proposed many times: to combine structured prior distributions on latent variables with complex dependencies implemented using deep neural networks~\citep[e.g.]{khan2017conjugate,archer2015black,klushyn2019learning,kosiorek2018sequential,burgess2019monet,greff2017neural,van2018relational,greff2019multi}.
The inference idea is more unique. 
Like standard variational autoencoders~\citep{kingma2013auto}, the SVAE uses a recognition network to aid in inferring the latent variables.
But rather than producing the full posterior, the SVAE recognition network only produces conjugate potentials.
The conjugate potentials are combined with the prior, and the posterior is computed using classical message passing algorithms~\citep{wainwright2008graphical}.

For example, consider an SVAE with a linear Gaussian dynamical system prior (LDS-SVAE) for sequential data. 
The generative model specifies a joint distribution over a sequence of latent variables~$x_{1:T}$ and emissions~$y_{1:T}$,
\begin{align*}
    p_{\theta,\gamma}(x_{1:T}, y_{1:T}) 
    &= p_\theta(x_1) \prod_{t=2}^T p_\theta(x_t \mid x_{t-1}) \prod_{t=1}^T p_\gamma(y_t \mid x_t),
\end{align*}
where $\theta$ are the prior parameters and $\gamma$ are the parameters of the decoder.
The prior is a linear dynamical system,
\begin{align*}
\begin{split}
    p_\theta(x_1) &= \mathcal{N}(x_1; \mu_1, Q_1), \\
    p_\theta(x_t \mid x_{t-1}) &= \mathcal{N}(x_t; A x_{t-1}, Q),
\end{split}
\end{align*}
where $\theta = (\mu_1, Q_1, A, Q)$.
However, the emission distribution, ${p_\gamma(y_t \mid x_t) = \mathcal{N}(y_t; f_\gamma(x_t), R)}$,
may be nonlinear, since $f_\gamma(\cdot)$ can be parameterized with a neural network with weights $\gamma$.

The SVAE parameters are learned by maximizing an evidence lower bound (ELBO),
\begin{align*}
    \mathcal{L}(\theta, \gamma, \phi) &= 
    \mathbb{E}_{q_{\theta,\phi}(x_{1:T}; y_{1:T})}\Big[ \log p_{\theta,\gamma}(x_{1:T}, y_{1:T})\\
    &\hspace{2em} -\log q_{\theta, \phi}(x_{1:T}; y_{1:T})\Big] 
    \leq \log p_{\theta,\gamma}(y_{1:T}),
\end{align*}
where~$q_{\theta,\phi}(x_{1:T}; y_{1:T})$ is a variational posterior distribution that is determined by both the variational parameters $\phi$ and the prior parameters $\theta$.

The SVAE defines the variational posterior in a clever way. 
It defines the posterior as the implicit solution to a \textit{surrogate variational inference problem},
\begin{align*}
    q_{\theta, \phi}(x_{1:T}; y_{1:T}) 
    &= \argmin_{\tilde{q} \in \mathcal{Q}} 
    \KL\left(\tilde{q}(x_{1:T})\,||\,\tilde{p}_{\theta, \phi}(x_{1:T} ; y_{1:T}) \right).
\end{align*}
The target,~$\tilde{p}_{\theta, \phi}(x_{1:T} ; y_{1:T})$, combines the prior with conjugate potentials,~$\psi_\phi(x_t; y_t)$, from the recognition network, 
\begin{align*}
\tilde{p}_{\theta,\phi}(x_{1:T} ; y_{1:T}) 
&\propto p_\theta(x_1) \prod_{t=2}^T p_\theta(x_t \mid x_{t-1}) \prod_{t=1}^T \psi_\phi(x_t; y_t), \\
\psi_\phi(x_t; y_t) &= \mathcal{N}(x_t; m_\phi(y_t), V_\phi(y_t)).
\end{align*}
Here, $m_\phi(y_t)$ and $V_\phi(y_t)$ are the mean and covariance of a Gaussian potential output by a neural network with weights~$\phi$. 

The key is that the surrogate inference problem is easy by design. 
Since the target is the product of a linear Gaussian prior and conjugate Gaussian potentials, the surrogate problem admits an exact solution via the Kalman smoother, a classic message passing algorithm. 
Figure~\ref{fig:svae_inference} illustrates this structured-exploiting amortized inference approach.

The SVAE exploits structure in the prior distribution by incorporating it into the target and only learning to produce conjugate potentials.
If the conjugate potentials are good approximations to the likelihoods, $p_\gamma(y_t \mid x_t)$, the solution to the surrogate problem will approximate the true posterior. 

The principal challenge in implementing an SVAE --- and we suspect the main reason why it has not been more widely adopted --- is that maximizing the ELBO requires computing gradients of the implicitly defined variational posterior with respect to the prior and variational parameters. 
In the case of the linear Gaussian prior above, this amounts to back-propagating gradients through a Kalman smoother. 
More generally, it may require back-propagating gradients through inference algorithms like coordinate-ascent variational inference~\citep{blei2017variational}. 

\subsection{Alternative Inference Approaches for Sequential VAEs}
A simpler and more commonly used approach is to have the recognition network output the full posterior directly.
A canonical example of this approach is the deep Kalman filter (DKF)~\citep{krishnan2015deep}. 
DKFs are deep generative models for sequential data, allowing for nonlinear latent variable dynamics and emissions. 
For posterior inference, the key difference is that in a DKF, the encoder is a bidirectional recurrent neural network (RNN) that directly maps the data $y_{1:T}$ to a variational posterior,
\begin{align*}
q_\phi(x_{1:T};y_{1:T})
=\prod_{t=1}^T \N(x_t; m_{\phi,t}, V_{\phi,t}),
\end{align*}
where $\{m_{\phi,t}, V_{\phi,t}\}_{t=1}^T$ are the means and covariances output by a bidirectional RNN with weights $\phi$ running over the data $y_{1:T}$. We call this the \textbf{RNN-MF} posterior family (fig.~\ref{fig:dkf_inference}), since the RNN outputs a mean-field (i.e.,~independent) posterior.

It is straightforward to extend this family to include temporal dependencies,
\begin{align*}
&q_\phi(x_{1:T};y_{1:T}) = \\
&\hspace{2em} \N(x_1; m_{\phi,1}, V_{\phi,1}) 
\prod^T_{t=2}\N(x_t; A_{\phi,t} x_{t-1} + m_{\phi,t}, V_{\phi,t}),
\end{align*}
where $\{m_{\phi,t}, A_{\phi,t}, V_{\phi,t}\}_{t=1}^T$ are outputs of a bidirectional RNN with weights $\phi$, and their dependence on the input data $y_{1:T}$ is omitted. 
We call this the \textbf{RNN-AR-L} posterior family (fig.~\ref{fig:cdkf_inference}), since it extends the RNN-MF family with linear autoregressive dependencies. 
Though we do not know of examples where this family has been used, it is of theoretical interest for our experiments since it contains the true posterior for linear Gaussian state space models.

A more commonly used posterior approximation is,
\begin{align*}
&q_\phi(x_{1:T};y_{1:T}) = \N(x_1; m_{\phi,1}, V_{\phi,1}) \\
&\hspace{4em} \times 
\prod^T_{t=2}\N(x_t; g_\phi(x_{1:t-1}, u_{\phi,t}), Q_\phi(x_{1:t-1}, u_{\phi,t})),
\end{align*}
where $g$ and $Q$ are neural networks, and $\{u_{\phi,t}\}_{t=1}^T$ are the outputs of a bidirectional RNN. 
This \textbf{RNN-AR-NL} posterior family (fig.~\ref{fig:planet_inference}) extends the one above by allowing \textit{nonlinear} autoregressive dependencies. 
For example, this style of posterior is used in LFADS~\citep{pandarinath2018inferring}, a sequential VAE for modeling neural spike train data. A similar posterior family is employed by PlaNet~\citep{hafner2019learning}, a deep architecture for model-based planning from image data.

Of course, the RNNs could be replaced with other architectures as well, like convolutional neural networks (CNNs)~\citep{bai2018empirical}, Transformers~\citep{vaswani2017attention}, or state space layers~\citep{gu2021efficiently, smith2022simplified} for sequence-to-sequence mapping. As another baseline, we consider a CNN recognition network with fixed-sized convolution kernels over the time dimension, paired with the linear Gaussian posterior. We call this the \textbf{CNN-AR-L} posterior family (fig.~\ref{fig:conv_inference}). While temporal CNNs are limited by the finite kernel size, they can still offer effective means of capturing temporal dependencies~\citep{bai2018empirical}.

\section{Modernizing the SVAE}
\label{section:svae}

We revisit the SVAE with three contributions that address the challenges that limited the original work:
\begin{itemize}
    \item We provide a JAX implementation that's modular and easy to use.
    \item We leverage parallel Kalman filtering and smoothing for significant speed-ups on parallel hardware.
    \item We propose a simple self-supervised scheme for learning latent dynamics that plays into the strengths of the SVAE in handling missing data. 
\end{itemize}

\subsection{Efficient JAX Implementation}
When the SVAE was first proposed, it did not see much practical usage because of its inherent complexity of implementation and the lack of hardware acceleration in the original implementation. With the advent of JAX~\citep{jax2018github}, a Python library allowing automated compilation for efficient gradient computation through arbitrary compositions of functions, the barriers to implementing sophisticated algorithms like those in the SVAE are lowered significantly.  Furthermore, the GPU and TPU support provided by JAX makes training more efficient. The JAX ecosystem provides libraries like JaxOpt~\citep{blondel2022efficient} for automatic \emph{implicit} differentiation, and Dynamax~\citep{dynamax2022github} for message passing in probabilistic state space models. In this work we provide a JAX implementation of SVAEs for sequential data, enabling fast and flexible model building and training.

\subsection{Parallel Message Passing}
In this section we introduce a strategy for efficiently parallelizing SVAE inference with a linear Gaussian dynamical system prior that can also be applied more generally. In the LDS-SVAE described above, the variational posterior is obtained by a Kalman smoother.
Standard implementations of the algorithm use a forward-backward recursion, with complexity that is linear in the length of the time series~\citep{sarkka2013bayesian}. However, this seemingly sequential algorithm can be efficiently parallelized by casting the process of forward filtering and backward smoothing/sampling as an associative operation on Gaussian distributions.  We can leverage the prefix sum algorithm for parallel computational span that is only \emph{logarithmic} in sequence length~\citep{sarkka2020temporal}. 

To illustrate the idea behind this, consider the following linear Gaussian chain as a simple case:
\begin{equation*}
    x_1 \sim \N(b_1, Q_1),\quad x_t \mid x_{t-1} \sim \N (A_t x_{t-1} + b_t, Q_t).
\end{equation*}
To compute $p(x_T)$, note that the linear Gaussian model implies Gaussian marginals, $p(x_t)=\N(x_t; \mu_t, \Sigma_t)$ for some mean $\mu_t$ and covariance $\Sigma_t$. The naive idea would be to start from $\mu_1=b_1$, $\Sigma_1=Q_1$ and iterate over $t$, marginalizing the $x_t$'s one at a time:
\begin{equation*}
\mu_t = A_t\mu_{t-1} + b_t, \quad \Sigma_t = A_t\Sigma_{t-1}A_t^\top+Q_t,
\end{equation*}
which seems like an inherently sequential process. Writing out another iteration, we see more structure to the problem:
\begin{align*}
    \mu_{t+1} &= A_{t+1}\mu_{t} + b_{t+1}\\
    &=A_{[t,t+1]}\mu_{t-1}+b_{[t,t+1]}\\
    \Sigma_{t+1} &= A_{t+1}\Sigma_{t}A_{t+1}^\top+Q_{t+1}\\
    &=A_{[t,t+1]}\Sigma_{t-1}A_{[t,t+1]}^\top+Q_{[t,t+1]},
\end{align*}
where $A_{[t,t+1]} = A_{t+1} A_t$, $b_{[t,t+1]} = A_{t+1}b_t+b_{t+1}$, and $Q_{[t,t+1]} = A_{t+1}Q_{t}A_{t+1}^\top+Q_{t+1}$.

The crucial observation is that this process of marginalization is \emph{closed} and \emph{associative}: the composition of two linear Gaussian conditional distributions gives another linear Gaussian conditional distribution, and the order of compositions does not affect the result. Instead of iterating on the means and covariances, we keep track of the parameters of the conditional distribution $a_t\equiv(A_t, b_t, Q_t)$, and define the following \textit{associative operator} $\otimes$ on them:
\begin{equation*}
\begin{split}
    (A_{i}, b_{i}, Q_{i}) &\otimes (A_{j}, b_{j}, Q_{j})\equiv (A_{[ij]}, b_{[ij]}, Q_{[ij]}) \\
    \text{where}&\;
    \begin{cases}
     A_{[ij]}=A_{j}A_{i}\\ 
     b_{[ij]}=A_{j}b_{i}+b_{j}\\
     Q_{[ij]}=A_{j}Q_{i}A_{j}^\top+Q_{j}
    \end{cases}
\end{split}
\end{equation*}
Notice that here the subscripts represent subsequences of $(1,2,\dots,T)$, and we used $[ij]$ to represent a concatenation of subsequences $i$ and $j$. It is easy to see that $\mu_t=b_{1:t}$ and $\Sigma_t=Q_{1:t}$. More importantly, the 
associative nature of the operation allows us to compute $a_{1:t}$ for all $t\in[1,T]$ in parallel, via the well-known prefix-sum algorithm~\citep{ladner1980parallel}.
Dynamax offers convenient JAX implementations of parallel Kalman filtering and smoothing~\citep{dynamax2022github}, enabling SVAE inference and training to have logarithmic parallel span. 
That is, the computation can be performed in $\mathcal{O}(\log T)$ time on a parallel machine.

\subsection{Self-supervised Learning and Missing Data}
\label{section:masking}

As observed in previous works, learning accurate dynamics from high-dimensional observations is a non-trivial task for latent variable generative models~\citep{karl2016deep,klushyn2021latent}. Practitioners often explicitly define losses based on current model extrapolations and true future observations. While this is an effective strategy, it is also a departure from the standard generative modeling approach. 

To train a sequential generative model to learn accurate dynamics, we use a simple self-supervised training method inspired by~\citep{kenton2019bert}. During each training iteration, we apply a random mask to the input data that zeros out a continuous subsequence of observations. Since the model is required to reconstruct the original unmasked data, it must rely on its learned dynamics to fill in the missing observations.

The SVAE has a unique advantage with this self-supervised scheme due to its structured approach to inference. Since each observation produces a separate conjugate potential, we can apply the mask by simply dropping the corresponding potentials (or equivalently, taking the potential covariance to infinity). This is a principled way to accommodate missing data in the SVAE and force it to rely on prior dynamics.

In practice, we found that masking out as much as 40\% of the input sequence encourages learning of the dynamics. We train all SVAE instances with masks applied throughout training, and we will show it yields substantial increase in predictive performance. For RNN and CNN-based models, however, removing such a large chunk of data at the beginning of training can be disastrous, resulting in posterior collapse~\citep{lucas2019understanding}. 

\section{Results}
\label{section:results}



\begin{figure}
\centering
    \begin{subfigure}{.38\textwidth}
        \centering
    \includegraphics[width=\linewidth]{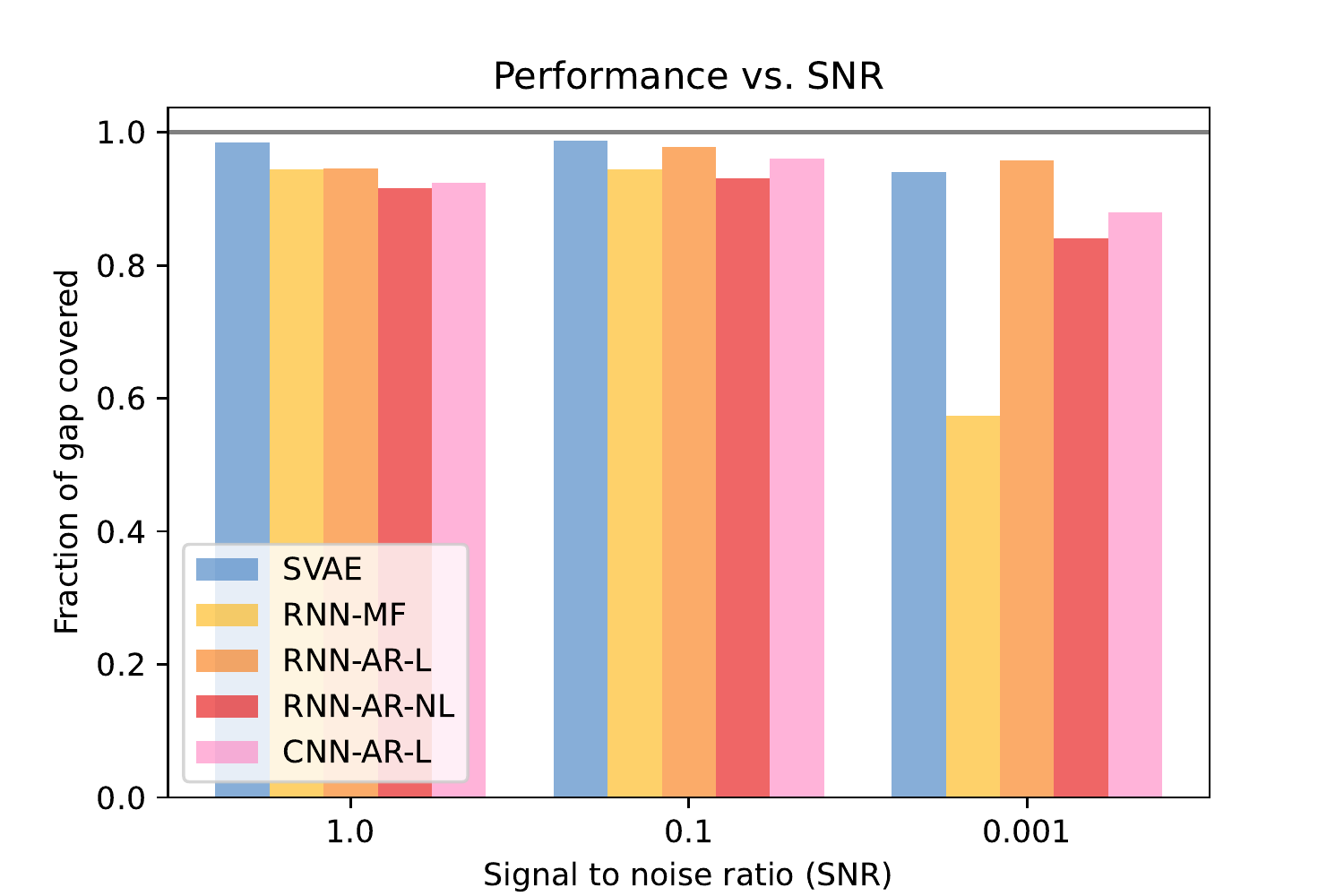}
    \end{subfigure}
    
    \begin{subfigure}{.38\textwidth}
        \centering
    \includegraphics[width=\linewidth]{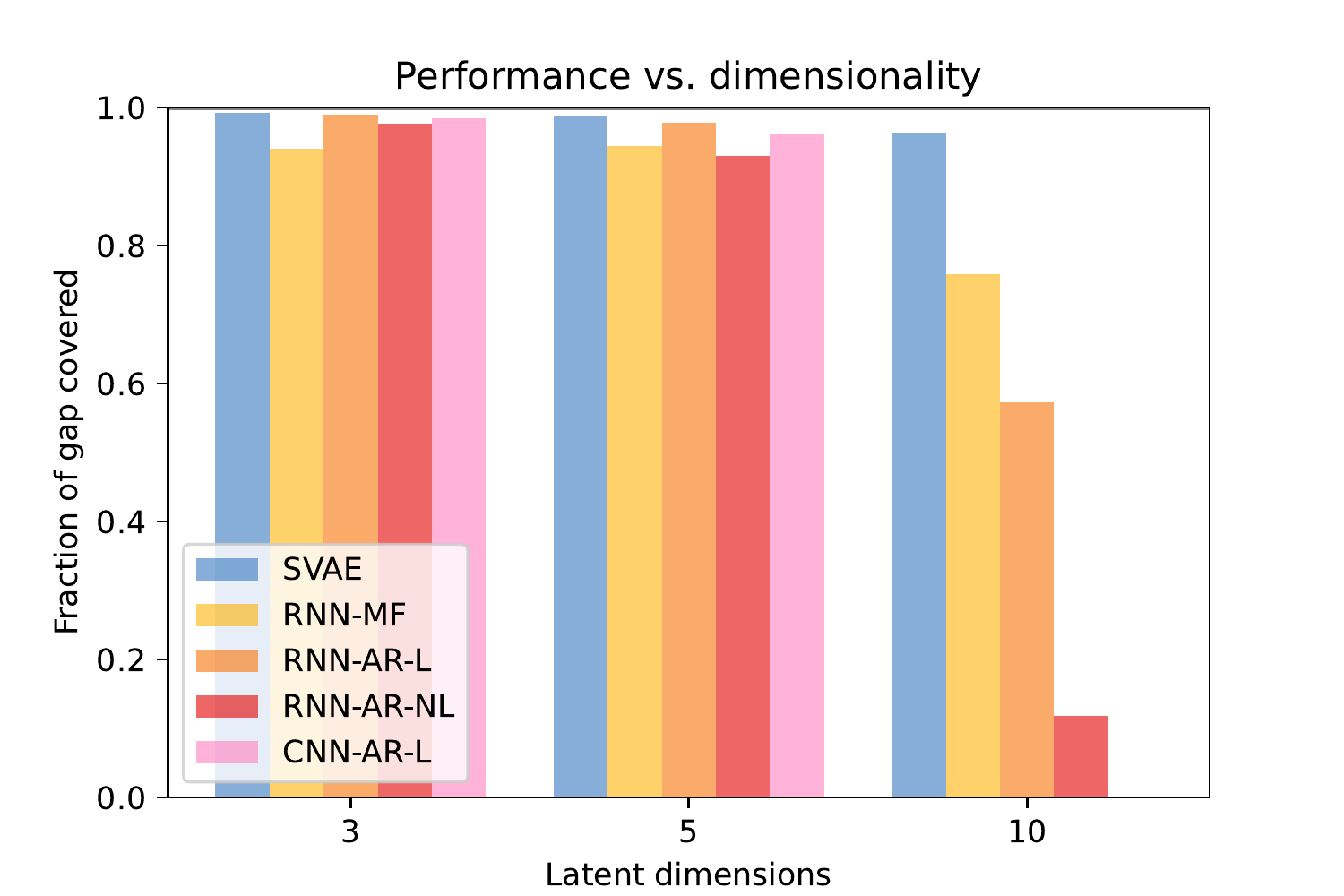}
    \end{subfigure}
    
    \caption{Top results on the LDS dataset. Performance between different methods is on par for most regimes, with the SVAE consistently achieving the best performance. Latent dimension size and SNR are 5 and 0.1 for the top and bottom plots respectively.}
    \label{fig:lds_results}
\end{figure}

\subsection{Linear Dynamical System Dataset}
\label{section:lds_results}
We first test all of the inference architectures from Section \ref{section:background} on toy datasets sampled from randomly generated linear dynamical systems. We test accuracy of the learned models under different state and observation dimensions, as well as different signal-to-noise (SNR) ratios. 

\paragraph{Dataset} We consider the following linear dynamical system with stationary dynamics:
\begin{align*}
x_1 &\sim \N(0, Q_1),\\
x_t \mid x_{t-1} &\sim \N(Ax_{t-1}+b, Q),\\
y_t \mid x_t &\sim \N(Cx_t, R),
\end{align*}
for $t\in[1,T]$, where $x_t\in\R^{D}$ and $y_t\in\R^{N}$. Furthermore, we sample $A$ as a rotation around a random axis in the $D$-dimensional latent space and set $Q=qI$, $R=rI$ to be diagonal covariance matrices.

We test the different recognition network architectures under three different noise scale combinations ${(q, r)\in \{(0.1, 0.1), (0.1, 1.0), (0.01, 10.0)\}}$ and three dimensionality settings $(D, N)\in \{(3, 5), (5, 10), (10, 20)\}$. We define the SNR as $q/r$.

For all of the parameter settings, the sequence length is 200 frames and we sample 100 sequences as the training data. 

\paragraph{Model architecture} For the SVAE, we use one linear layer with two linear readouts for the mean and covariance of the potential, since we know that the optimal recognition potential for LDS data is linearly related to the inputs. For the RNN models, we use various different recurrent state sizes $H\in\{10, 20, 30\}$, and we also add one to two hidden layers with ReLU nonlinearity to the output heads $A_{\phi,t}$, $b_{\phi,t}$ and $V_{\phi,t}$. For RNN-AR-NL we implement the nonlinear conditional distribution as a gated recurrent unit (GRU) cell. For CNN-AR-L, we use a 3-layer architecture with 32 features in each layer, and convolution kernels in the time dimension of sizes up to 50 timesteps. We use the same decoder network for all of the models, which is a one layer linear network with linear readouts for the output mean and covariance.

\begin{figure}
    \centering
    \includegraphics[width=.8\linewidth]{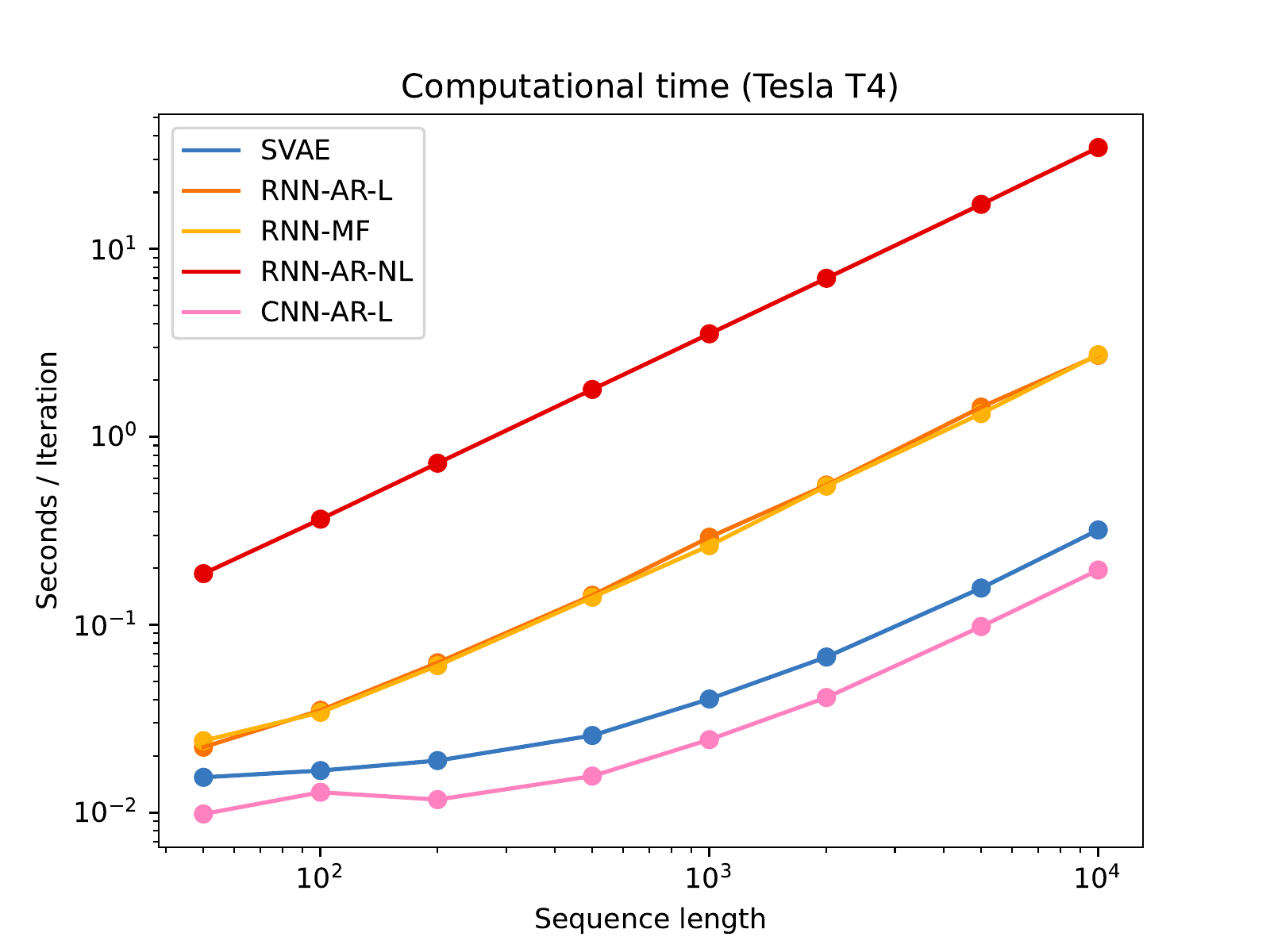}
    \caption{Wallclock time per iteration of the different methods on a Tesla T4 GPU. The SVAE with parallel scan achieves nearly ten-fold speedup to the faster RNN methods.}
    \label{fig:run_time}
\end{figure}

\paragraph{Null model} To have a basis for comparison for the SVAE and RNN-based methods, we establish a static baseline model:
\begin{equation*}
    p_{\mathsf{st}}(x_{1:T}, y_{1:T}) = \prod_tp_{\hat{\theta}_t}(x_t)p_{\hat{\gamma}_t}(y_t\mid x_t),
\end{equation*}
where $p_{\hat{\theta}_t}(x_t)=\N(x_t;0, \hat{Q}_t)$ is the marginal distribution of $x_t$ and $p_{\hat{\gamma}_t}(y_t| x_t)=\N(y_t;Cx_t,R)$ is the emissions distribution under the true model. Since both terms are linear and Gaussian, the marginal distribution of the static model is $p_{\mathsf{st}}(y_{1:T})=\prod_{t=1}^T \N(y_t; 0, \hat{Q}+CRC^\top)$. This baseline is the best ELBO achievable by a ``static'' model, e.g. a VAE that treats each of the latent variables as independent across time. Intuitively, the noisier the observations, the harder it is for the models to learn the dynamics. Therefore, we expect the learned models to outperform the static baseline by a smaller margin in these regimes.

\paragraph{Results} We perform hyperparameter search for all of the methods and show the best ELBO achieved in figure~\ref{fig:lds_results}. We normalize the ELBO by looking at the percentage of the gap covered between the marginal data log likelihood under the null model $p_{\mathrm{st}}(y_{1:T})$ and the true model $p_{\mathrm{true}}(y_{1:T})$. Overall, the SVAE is the most consistent and achieves the highest ELBO. Unsurprisingly, we notice that low SNR and high dimensionality negatively impact model performance, resulting in decreased performance for all methods, with the SVAE being the least impacted. Note that these trends also scale higher dimensions, which we explore  in Appendix~\ref{app:high_d}.

\subsection{Computational Efficiency of SVAE with Parallel Kalman Filtering}
Here we examine the practical benefits of the parallel Kalman filter in the SVAE. We run all of the methods on synthetic LDS data with three-dimensional latent states and five-dimensional observations. We scale up the sequence length and measure the wallclock time it takes for each method to complete an iteration over a batch of ten sequences on a Tesla T4 GPU. For the RNN and CNN-based methods, we use the same architecture as Section \ref{section:lds_results} with ten-dimensional latent states for the RNNs and a kernel size of 10 for the CNN.

The results are shown in figure~\ref{fig:run_time}. While we do not achieve the theoretical efficiency of $\mathcal{O}(\log T)$ in sequence length $T$ due to practical limits on the amount of parallel resources units available, the SVAE still obtained almost ten-fold speedup over the mean-field and autoregressive RNN methods. Unsurprisingly, the nonlinear autoregressive RNN is the slowest, with an additional RNN for sequential sampling from the posterior alongside the sequential BiRNN inference network. The CNN-AR-L is slightly faster than the SVAE but follows the same trend, however with finite kernel size it cannot capture the same temporal dependencies.

\subsection{Synthetic Pendulum Dataset}

\begin{figure*}
\centering
    \begin{subfigure}{.45\textwidth}
    \centering
    \includegraphics[width=\linewidth]{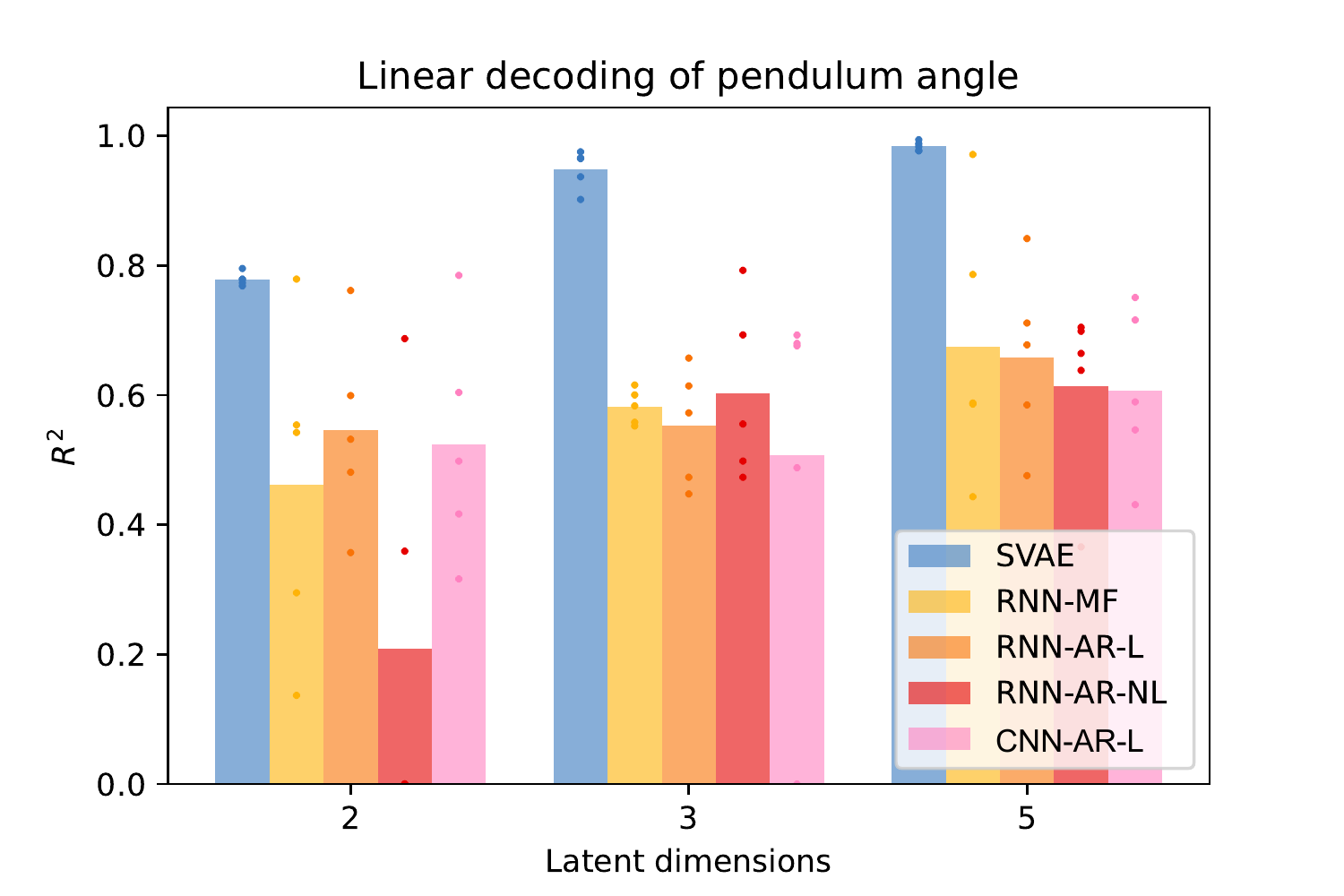}
    \end{subfigure}
    \begin{subfigure}{.45\textwidth}
    \centering
    \includegraphics[width=\linewidth]{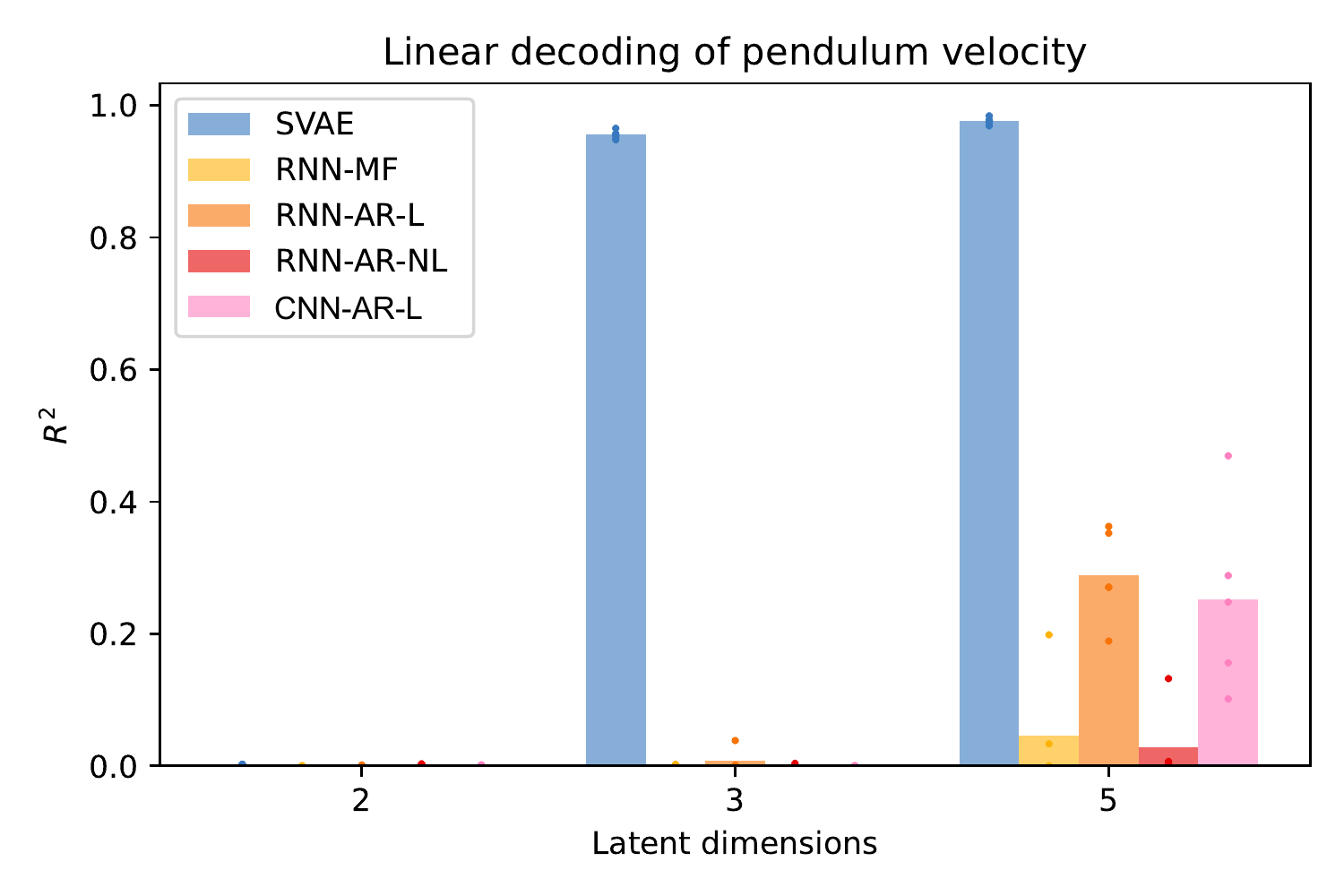}
    \end{subfigure}
    \caption{SVAE learns representations of pendulum data that are informative about the dynamics. We run linear regression from the latent representations to the true pendulum angles (left) and angular velocities (right) and report the $R^2$ values. Notably, the SVAE is the only method that learns a representation of angular velocity with 3 dimensional latents.}
    \label{fig:linear_decode}
\end{figure*}

It is perhaps no surprise that the SVAE performs well on data generated from the same prior class with linear emissions. In general, we are more interested in the value of SVAEs in more complex settings, where the emissions are high dimensional and nonlinear. Furthermore, it would be interesting to see the SVAE perform in a regime where the dynamics are \emph{nonlinear}.

We apply the SVAE and the RNN methods to the synthetic pendulum dataset adapted from \citet{schirmer2022modeling}. A $24\times24$ pixel movie of a swinging pendulum is rendered with noise, and the task is to learn the underlying low-dimensional dynamics in an unsupervised fashion. We take regularly spaced observations of the pendulum and apply uniform Gaussian noise stationary across time to each pixel. Since the true physical state (angle and angular momentum) of the pendulum can be described in a two-dimensional state space, we use models with small latent dimensionality $D\in(2, 3, 5)$. It is important to note that for most of this dataset the small-angle approximation does not apply, and therefore the underlying dynamics are nonlinear.

One challenge in learning dynamics is that angular velocity information is missing from the individual frames. Since the pendulum observations exist in a one-dimensional manifold, the models do not need dynamics information to be able to reconstruct the observations. To encourage the models to recover the underlying dynamics, we apply the self-supervised masking introduced in Section \ref{section:masking} to all of the models. We found that RNN-based recognition networks failed to learn accurate models when significant masking was used in early stages of training; thus, we gave these architectures full observations for the first 10-100 epochs before applying the masks. For the SVAE, we found it trained reliably even with masks applied from the start. This is likely due to the inference algorithm of the SVAE, which allows it to handle missing data in a principled way. We compare the results of the SVAE, RNN-AR-L and RNN-AR-NL on 100 sequences of pendulum renderings, each with 100 frames. We pick the top five runs in ELBO on the test set and analyze the learned models. The achieved ELBO values with and without the self supervised masking is reported in appendix~\ref{app:ablation}. The SVAE achieves comparable ELBO values to other methods, but its learned models yield better representations of the data dynamics, as we show next.

\paragraph{SVAE learns linearly decodable representations} We find that the true physical states of the pendulum can be linearly decoded from the latent states inferred by the SVAE. Figure~\ref{fig:linear_decode} shows $R^2$ values of a linear regression from the learned latents to true angles and angular velocities. Across different latent sizes, the SVAE admits representations that are more linearly related to the true physical states. Note that none of the models are able to encode information about angular velocity given two-dimensional latents, while the SVAE is able to explain almost all of the variance in angular velocity given an additional dimension.

\begin{figure}
\centering
    \begin{subfigure}{.45\textwidth}
    \centering
    \includegraphics[width=\linewidth]{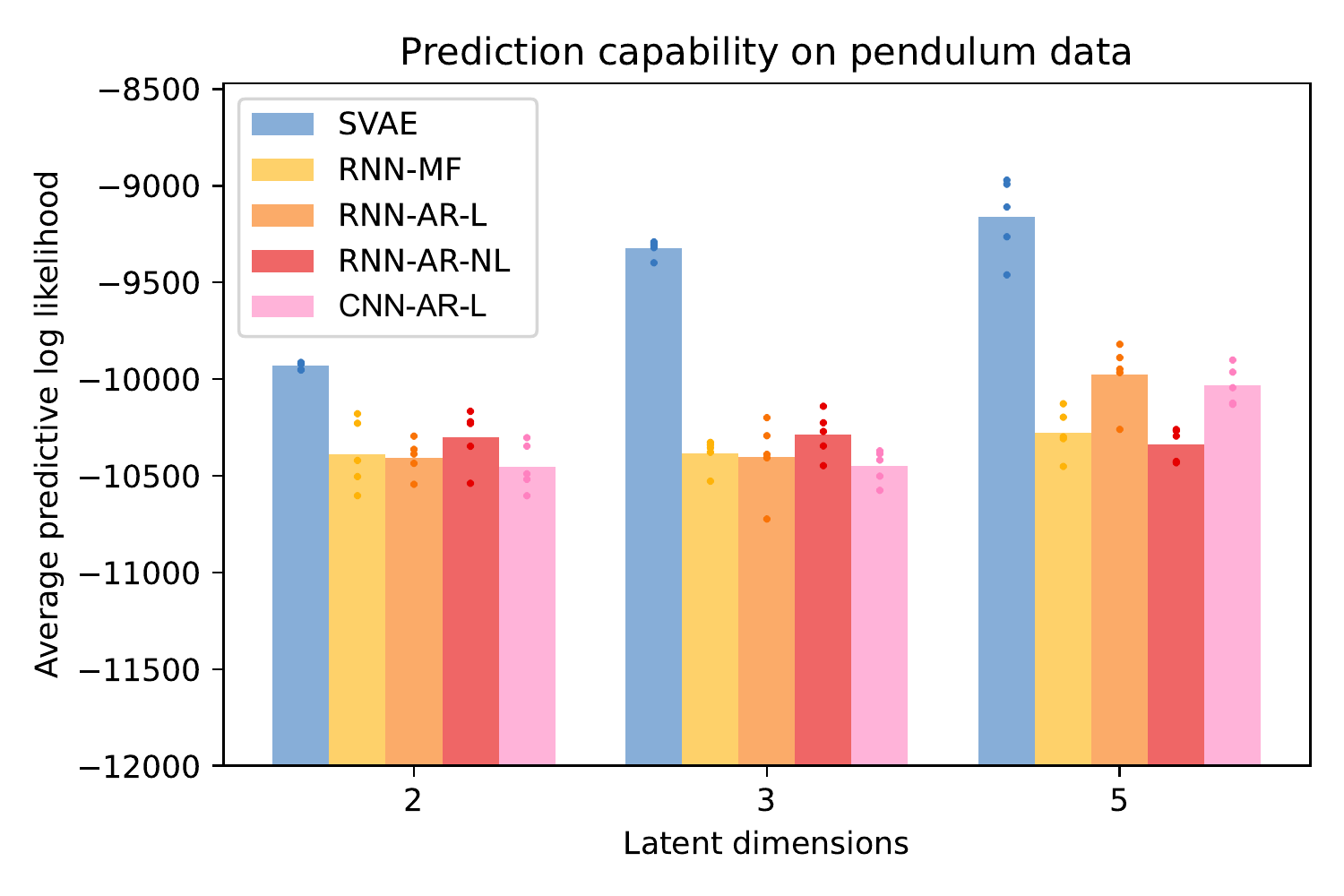}
    \end{subfigure}
    \begin{subfigure}{.45\textwidth}
    \centering
    \includegraphics[width=\linewidth]{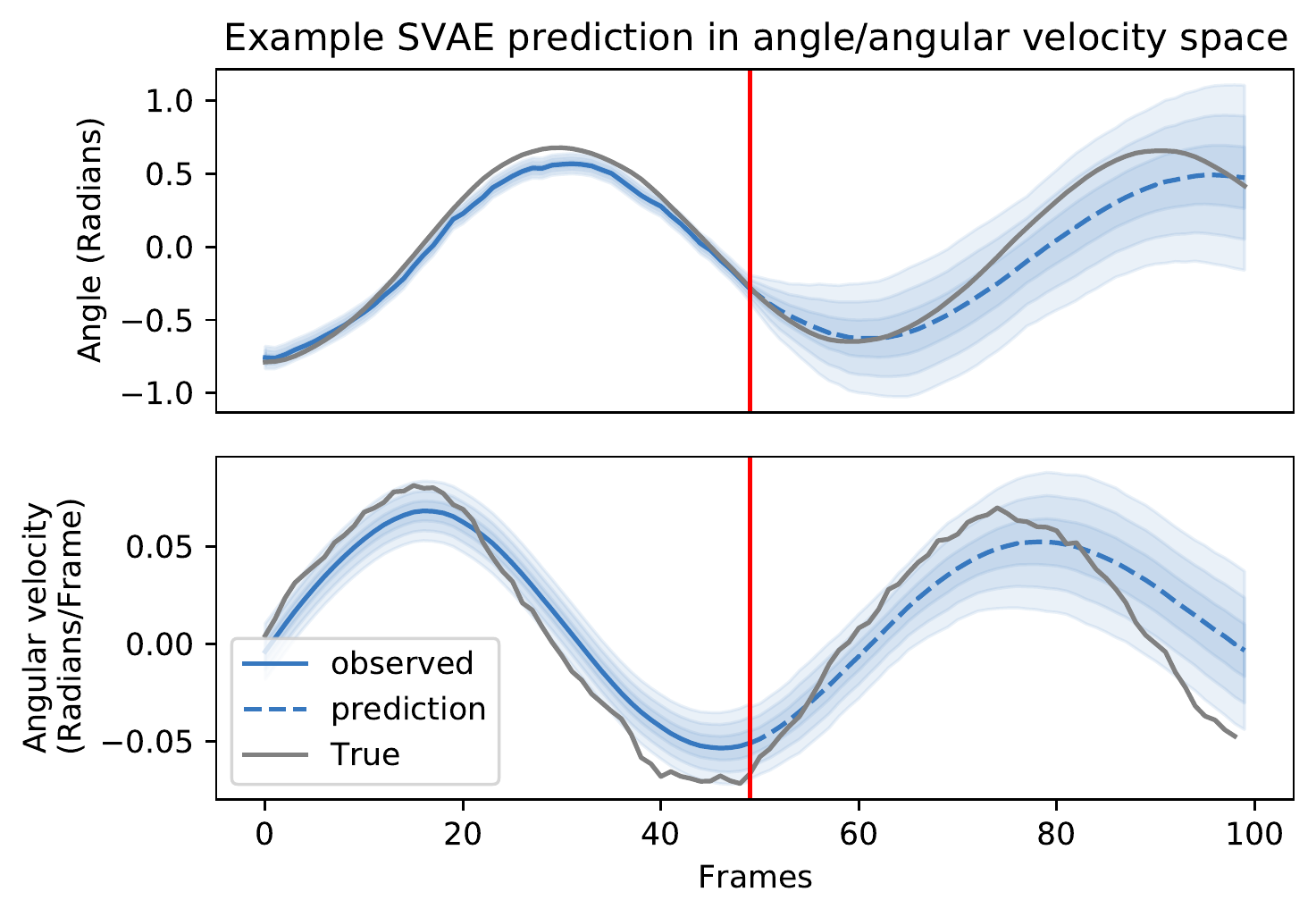}
    
    \end{subfigure}
    \caption{SVAE yields better predictive performance. We sample many latent trajectories from the learned dynamics and estimate the marginal log likelihood of the future 50 frames give the past 50 frames (top). Given the linear regression weights, we can make predictions in the phase space of the pendulum (bottom). Each shade in the plot corresponds to one standard deviation.}
    \label{fig:prediction_ll}
    \label{fig:phase_space_prediction}
\end{figure}

\begin{figure*}[ht!]
\centering
    \includegraphics[trim={0cm 4cm 5cm 5cm},width=1\linewidth]{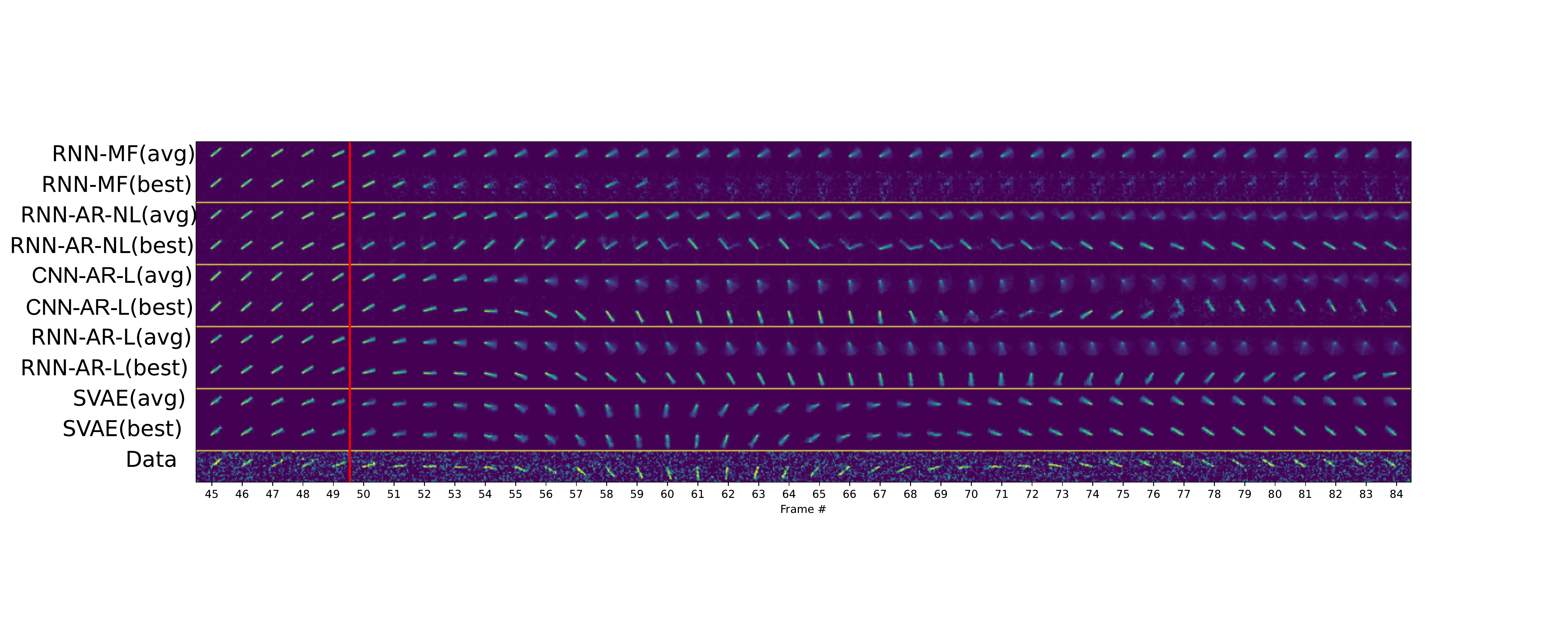}
    \caption{Sample model predictions for pendulum data in image space. For all of the models, we show both the average prediction (avg) and the best prediction (best) among the 200 sample trajectories. We see that the SVAE alone predicts pendulum angles that are concentrated around the true angles.}
    \label{fig:image_prediction}
\end{figure*}

\paragraph{Model performance on prediction} Given the first 50 frames of a test sequence, We forecast latent states with the learned dynamics model $p_\theta(x_{t+1}\mid x_t)$ and estimate the log marginal likelihood of true data $\log p(y_{51:100}\mid y_{1:50})$ by approximating the integral over future latent states with Monte Carlo. Across different latent space sizes, the SVAE yields higher likelihoods for future data (fig.~\ref{fig:prediction_ll}). Given the linear decodability of the SVAE latents, we also visualize an example predicted trajectory of angles and angular velocities in figure~\ref{fig:phase_space_prediction}. 

We show sample predictions in the image space in figure~\ref{fig:image_prediction}. To show how the models represent uncertainty in their predictions, we sample 200 latent trajectories $\{x^{(i)}_{1:T}\}_{i=1}^{200}$ of length $T=100$. We train on the first 50 time steps and forecast the last 50. Fig.~\ref{fig:image_prediction} shows both the most accurate predictions and the average predicted frames. While the best prediction from the RNN-AR-L are sharp and reasonably accurate, the average prediction quickly becomes diffuse. The SVAE, on the other hand, is able to maintain its spread over angular positions throughout the sequence, demonstrating the accuracy of its learned dynamics model. 

This pendulum example would not be possible with the existing, pure Python SVAE implementation, since it requires convolutional neural networks for generating the recognition potentials and implementing the decoder. With our JAX implementation, the model can be trained on a single GPU in tens of minutes. The promising results on this image forecasting task suggest that the SVAE can be a valuable model for more complex, high-dimensional time series as well.

\section{Related Works}
\label{section:related_work}
In recent years, there has been a plethora of work on deep sequential LVMs, with applications ranging from speech modeling, and video compression to prediction and planning in real and simulated environments~\citep{chung2015recurrent,li2018disentangled, marino2018general,hafner2019learning,babaeizadeh2018stochastic,karl2016deep,saxena2021clockwork}. We focus on the LDS-SVAE since it is a simple yet interesting model that demonstrates the advantages of structure-exploiting inference for complex, high-dimensional datasets. It is worth noting that the SVAE also allows for more complex and expressive PGM priors such as the switching linear dynamical system (SLDS)~\citep{murphy1998switching,fox2008nonparametric} and recurrent switching linear dynamical system (rSLDS)~\citep{linderman2017bayesian} for the purposes of analyzing sequential data. 

\paragraph{Amortization gap} Amortized inference with inflexible encoder networks results in significant \textit{amortization gaps} --- the difference between the ELBO and the true marginal likelihood that arises from the recognition network outputting suboptimal variational parameters~\citep{cremer2018inference,marino2018general}. A failure mode that often accompanies this amortization gap is the posterior collapse, which happens when the model learns to ignore the latents completely~\citep{alemi2018fixing,lucas2019don}. \citet{turner+sahani:2011a} studied the effects of approximation gap on model learning, and more recent work has considered this issue in the context of amortized variational inference~\citep{cremer2018inference,shu2018amortized,he2019lagging}. The SVAE addresses this gap by incorporating the prior into the implicit definition of the posterior distribution.

\paragraph{Structure-exploiting recognition networks} Some inference frameworks try to address the amortization gap issue by having the variational posterior depend on the prior parameters~\citep{marino2018general,johnson2016composing,lin2018variational,tomczak2018vae}. Amortized variational filtering (AVF)~\citep{marino2018general} uses inner-loop gradient updates to correct the variational posterior which accounts for the changing prior in an expectation-maximization-like fashion. Other approaches like the SVAE use structured exponential family priors and variational posteriors computed with an inner structured inference step involving the prior, offloading computation from the inference network. In this work we focus on the SVAE as one representative of structure-exploiting amortized inference and demonstrate how it performs more accurate inference and supports better parameter estimation over its non-structured counterparts. Interestingly, despite having the capacity to learn the true posterior in our linear dynamical systems experiments, we found that the RNN-based approaches failed to attain the true marginal log likelihood. In essence, it appears hard for RNNs to learn to perform Kalman smoothing with noisy, high-dimensional observations in the small data regimes that we consider.

\paragraph{Sequence modeling beyond RNNs} In the past few years there has been a burst of new model and architectures that tackles sequence modeling. Transformers~\citep{vaswani2017attention, kenton2019bert,brown2020language} have become the new standard in language modeling, outperforming RNNs. For very long sequences, structured state space model for sequences (S4)~\citep{gu2021efficiently} and its variants~\citep{smith2022simplified,hasani2022liquid,gupta2022diagonal} are becoming more prominent in solving hard sequence tasks with high parallel efficiency. Notably,~\citet{zhou2022deep} uses S4 in place of the prior, recognition and decoder components of a sequential VAE. Although we included the CNN as a simple alternative to the RNN-based models, it would be natural to follow up with comparisons between SVAEs and these new types of sequence modeling techniques that have more complex architectures and stronger modeling power. 
\section{Conclusion}

We revisit the structured variational autoencoder, a structure-exploiting amortized variational inference framework that combines the flexibility of neural networks with the modeling benefits of probabilistic graphical models. Our JAX implementation runs efficiently on modern hardware and leverages parallel Kalman filtering to achieve an order of magnitude speed-up over RNN-based unstructured inference methods.

We showed that the SVAE with linear Gaussian dynamics yields competitive performance on synthetic datasets. Not only does it perform fast and accurate inference in the case of well-specified models, it also yields superior predictive performance for high-dimensional observations with nonlinear underlying dynamics. With its advantage for handling missing data, we use self-supervised masking to encourage learning of dynamics. Compared to its RNN and CNN-based counterparts, the SVAE learns representations that are informative about the dynamics, and we were able to linearly decode the true states of the system from its latent states. These results show much promise for the SVAE in deep generative modeling tasks.

\section*{Acknowledgements}

This work was supported by the Stanford Interdisciplinary Graduate Fellowship (SIGF), along with grants from the Simons Collaboration on the Global Brain (SCGB 697092), the NIH (U19NS113201, R01NS113119, and R01NS130789), the Sloan Foundation, and the Stanford Center for Human and Artificial Intelligence.
We thank Matt Johnson for his constructive feedback and insightful discussions with us on the paper. Additionally, we thank Blue Sheffer and Dieterich Lawson for their contribution in the early stages of the project, Andy Warrington and Jimmy Smith for their extensive feedback in the paper writing process, and the rest of the Linderman Lab for their support and feedback throughout the project.
\bibliographystyle{unsrtnat}
\bibliography{99_main}
\newpage
\appendix
\onecolumn

\section{Learning higher dimensional linear dynamics}
\label{app:high_d}
To investigate if the performance of the SVAE can scale to higher dimensions, we include results on learning higher dimensional linear dynamical systems in Table~\ref{tab:app_high_d}. For the data, we sample from 32 and 64 dimensional linear dynamical systems with 64 and 128 dimensional observations respectively. For the 128 dimensional dataset, we also sampled 10 times more data, since we expect the sample complexity of parameter estimation to be greater. We use the same architecture as in the LDS experiments in Section~\ref{section:lds_results}, but scaling up the hidden layer sizes to $64$ and $128$. We also use diagonal matrices for the dynamics noise covariance and recognition potentials for efficiency. None of the models achieve the true marginal likelihood in this regime, but the SVAE still outperforms the other methods considerably.


\begin{table}[h!]
\vskip 0.15in
\caption{Evidence lower bound (ELBO) on higher dimensional linear dynamical systems.}
\begin{center}
\begin{small}
\begin{sc}
\begin{tabular}{lc|ccr}
\toprule
Method & \multicolumn{2}{c}{ELBO} \\
& 32d & 64d \\
\midrule
True MLL & $\mathbf{-1.905}$ & $\mathbf{-2.039}$ \\
\midrule
SVAE & $\mathbf{-2.599}$ & $\mathbf{-3.419}$ \\
RNN-AR-L & -4.331 & -4.665 \\
RNN-MF & -4.317 & -4.781 \\
RNN-AR-NL & -4.697 & -6.395 \\
CNN-AR-L & -4.387 & -7.005 \\
\bottomrule
\end{tabular}
\end{sc}
\end{small}
\end{center}
\label{tab:app_high_d}
\vskip -0.1in
\end{table}

\section{Ablation studies for unsupervised masking}
\label{app:ablation}
Here we examine the effects of the unsupervised masking approach to learning dynamics. We provide results in validation ELBO and predictive log likelihoods for all of the methods with and without masks applied in Table~\ref{tab:app_ablation}. The results shown here correspond to models with 5 dimensional latent spaces, but the same trends appear across the different dimensionalities studied in the paper. It is clear that while applying the masks hurt the overall ELBO achieved, due to having less access to and more uncertainty in the training data, it is beneficial for learning the correct model for predicting the future.

\begin{table}[h!]
\vskip 0.15in
\caption{Ablation results for the unsupervised masking method.}
\begin{center}
\begin{small}
\begin{sc}
\begin{tabular}{lcccr}
\toprule
Method & ELBO & Prediction \\
\midrule
SVAE & -0.3015 & $\mathbf{-0.1558}$ \\
SVAE (no mask) & $\mathbf{-0.2763}$ & -0.1910 \\
\midrule
RNN-AR-L & -0.2886 & $\mathbf{-0.1717}$ \\
RNN-AR-L (no mask) & $\mathbf{-0.2757}$ & -0.1862 \\
\midrule
RNN-MF & -0.2906 & $\mathbf{-0.1815}$ \\
RNN-MF (no mask) & $\mathbf{-0.2761}$ & -0.1899 \\
\midrule
RNN-AR-NL & -0.2914 & $\mathbf{-0.1783}$ \\
RNN-AR-NL (no mask) & $\mathbf{-0.2779}$ & -0.1798 \\
\midrule
CNN-AR-L & -0.2902 & $\mathbf{-0.1719}$ \\
CNN-AR-L (no mask) & $\mathbf{-0.2796}$ & -0.1861 \\
\bottomrule
\end{tabular}
\end{sc}
\end{small}
\end{center}
\label{tab:app_ablation}
\vskip -0.1in
\end{table}

\end{document}